# Objective evaluation metrics for automatic classification of EEG events

**Saeedeh Ziyabari[1], Vinit Shah[1], Meysam Golmohammadi[2], Iyad Obeid[1] and Joseph Picone[1]**

[1] The Neural Engineering Data Consortium, Temple University 1947 North 12th Street, Philadelphia, Pennsylvania, 19122, USA.

[2] BioSignal Analytics, Inc., 3624 Market Street, Suite 5E, Philadelphia, Pennsylvania, 19104, USA.

**Abstract:** The evaluation of machine learning algorithms in biomedical fields for applications involving sequential data lacks standardization. Common quantitative scalar evaluation metrics such as sensitivity and specificity can often be misleading depending on the requirements of the application. Evaluation metrics must ultimately reflect the needs of users yet be sufficiently sensitive to guide algorithm development. Feedback from critical care clinicians who use automated event detection software in clinical applications has been overwhelmingly emphatic that a low false alarm rate, typically measured in units of the number of errors per 24 hours, is the single most important criterion for user acceptance. Though using a single metric is not often as insightful as examining performance over a range of operating conditions, there is a need for a single scalar figure of merit. In this paper, we discuss the deficiencies of existing metrics for a seizure detection task and propose several new metrics that offer a more balanced view of performance. We demonstrate these metrics on a seizure detection task based on the TUH EEG Corpus. We show that two promising metrics are a measure based on a concept borrowed from the spoken term detection literature, Actual Term-Weighted Value, and a new metric, Time-Aligned Event Scoring (TAES), that accounts for the temporal alignment of the hypothesis to the reference annotation. We also demonstrate that state of the art technology based on deep learning, though impressive in its performance, still needs significant improvement before it will meet very strict user acceptance guidelines.

## 1. Introduction

Electroencephalograms (EEGs) are the primary means by which physicians diagnose, evaluate and manage brain-related illnesses such as epilepsy, seizures and sleep disorders [1]. Automatic interpretation of EEGs has been extensively studied in the past decade [2]-[6]. However, even though many researchers report impressive levels of accuracy in publications, widespread adoption of commercial technology has yet to happen in clinical settings primarily due to the high false alarm (FA) rates of these systems [7][8][9]. In this paper, we investigate the gap in performance between research and commercial technology and discuss how these perceptions are influenced by a lack of a standardized scoring methodology.

There are in general two types of ways to evaluate machine learning technology: user acceptance testing [10][11] and objective performance metrics based on annotated reference data [12][13]. User acceptance testing is time-consuming and expensive. It has never been a practical way to guide technology development because algorithm developers need rapid turnaround times on evaluations. Hence evaluations using objective performance metrics, such as sensitivity and specificity, are common in the machine

**Corresponding Author:** Joseph Picone, The Neural Engineering Data Consortium, ENGR 703A, Temple University, 1947 North 12th Street, Philadelphia, Pennsylvania, 19122, USA, Tel: 215-204-4841, Fax: 215-204-5960, Email: joseph.picone@gmail.com.

learning field [14][15][16]. With this approach, it is very important to have a rich evaluation dataset and a performance metric that correlates well with user and application needs. The metric must have a certain level of granularity so that small differences in algorithms can be investigated and parameter optimizations can be evaluated. For example, in speech recognition applications, word error rate has been used for many years because it correlates well with user acceptance testing but provides the necessary level of granularity to guide technology development. Despite many years of research focused on finding better performance metrics [17][18], word error rate remains a valid metric for technology development and assessment.

Sequential pattern recognition applications, such as speech recognition, keyword search or EEG analysis, require additional considerations. Data, typically organized in files on a computer, are not simply assessed with an overall judgment (e.g., "did a seizure occur somewhere in this file?"). Instead, the locality of the hypothesis must be considered – to what extent did the start and end times of the hypothesis match the reference transcription. This is a complex issue since a hypothesis can partially overlap with the reference annotation, and a consistent mechanism for scoring such events must be adopted. Unfortunately, there is no such standardization in the EEG literature. For example, Wilson et al. [19] advocates using a term-based metric involving sensitivity and specificity. A term was defined as a connection of consecutive decisions from the same type of event. A hypothesis is counted as a true positive when it overlaps with one or more reference annotations. A false positive corresponds to an event in which a hypothesis annotation does not overlap with any of the reference annotations. Kelly et al. [20] recommends using a metric that measures sensitivity and FAs. A hypothesis is considered a true positive when the time of detection is within two minutes of the seizure onset. Otherwise it is considered a false positive. Baldassano et al. [21] uses an epoch-based metric that measures false positive and negative rates as well as latency. The development, evaluation and ranking of various machine learning approaches is highly dependent on the choice of a metric.

A large class of bioengineering problems, including seizure detection, involve prediction as well as classification. In prediction problems, we are often concerned with how far in advance of an event (or after the event has occurred) we can predict an outcome. Accuracy of prediction varies with latency, so this type of performance evaluation adds some complexity to the process. Winterhalder et al. [22] have studied this problem extensively and argue for a scoring based on long-term considerations. In this paper, we are not concerned with these types of prediction problems. We are focused mainly on assessing the accuracy of classification and assessing the proximity of these classifications to the actual event.

Therefore, we analyze several popular scoring metrics and discuss their strengths and weaknesses on sequential decoding problems. We introduce several alternatives, such as the Actual Term-Weighted Value [23][24] that have proven successful in other fields, and discuss their relevance to EEG applications. We present a comparison of performance for several systems using these metrics and discuss how this correlates with overall user acceptance.

## 2. Method

Researchers in biomedical fields typically report performance in terms of sensitivity and specificity [25]. In a two-class classification problem such as seizure detection, we can define four types of errors:

True Positives (TP): the number of 'positives' detected correctly
True Negatives (TN): the number of 'negatives' detected correctly
False Positives (FP): the number of 'negatives' detected as 'positives'
False Negatives (FN): the number of 'positives' detected as 'negatives'

Sensitivity (TP/(TP+FN)) and specificity (TN/(TN+FP)) are derived from these quantities. There are a large number of auxiliary measures that can be calculated from these four basic quantities that are used extensively in the literature. These are summarized concisely in [26]. For example, in information retrieval problems, systems are often evaluated using accuracy ((TP+TN)/(TP+FN+TN+FP)), precision (TP/(TP+FP)), recall (another term for sensitivity) and $F_1$ score $((2\cdot Precision\cdot Recall)/(Precision + Recall))$. However, none of these measures address the time scale on which the scoring must occur, which is critical in the interpretation of these measures for many real-time bioengineering applications.

In some applications, it is preferable to score every unit of time. With multichannel signals, such as EEGs, scoring for each channel for each unit of time might be appropriate since significant events such as seizures occur on a subset of the channels present in the signal. However, it is more common in the literature to simply score a summary decision per unit of time that is based on the per-channel inputs (e.g., a majority vote). We refer to this type of scoring as epoch-based [27][28]. An alternative, that is more common in speech and image recognition applications, is term-based [24][29], in which we consider the start and stop time of the event, and each event identified in the reference annotation is counted once. There are fundamental differences between the two conventions. For example, one event containing many epochs will count more heavily in an epoch-based scoring scenario. Epoch-based scoring generally weights the duration of an event more heavily since each unit of time is assessed independently.

Time-aligned scoring is essential to sequential decoding problems. But to implement such scoring in a meaningful way, there needs to be universal agreement on how to assess overlap between the reference and the hypothesis. For example, Figure 1 demonstrates a typical issue in scoring. The machine learning system correctly detected *5* seconds of a *10*-sec event. Essentially *50%* of the event is correctly detected, but how that is reflected in the scoring depends on the specific metric. Epoch-based scoring with an epoch duration of *1* sec would count *5* FN errors and *5* TP detections. Term-based scoring would potentially count this as a correct recognition depending on the way overlaps are scored.

Term-based metrics score on an event basis and do not count individual frames. A typical approach for calculating errors in term-based scoring is the Any-Overlap Method (OVLP) [30][31]. TPs are counted when the hypothesis overlaps with reference annotation. FPs correspond to situations in which the hypothesis does not overlap with the reference. The metric ignores the duration of the term in the reference annotation. In Figure 2, we demonstrate two extreme cases for which the OVLP metric fails. In each case, *90%* of the event is incorrectly scored. In example no. 1, the system does not detect approximately *9* seconds of a seizure event, while in example no. 2, the system incorrectly labels an additional *9* seconds of time as seizure. OVLP is considered a very permissive way of scoring, resulting in artificially high sensitivities. In Figure 2, the OVLP metric will score both examples as 100% TP.

It is very difficult to compare the performance of various systems when only two values are reported (e.g. sensitivity and specificity) and when the prior probabilities vary significantly (in seizure detection, the a priori probability of a seizure is very low, which means assessment of background events dominate the error calculations). Often a more holistic view is preferred, such as a Receiver Operating Characteristic (ROC) [15] or a Detection Error Trade-off (DET) curve [16]. An ROC curve displays the TP rate as a function of the FP rate while a DET curve displays the FN rate as a function of the FP rate. When a single metric is preferred, the area under an ROC curve (AUC) [32][33] is also an effective way of comparing the performance. A random guessing approach to classification will give an AUC of *0.5* while a perfect classifier will give an AUC of *1.0*.

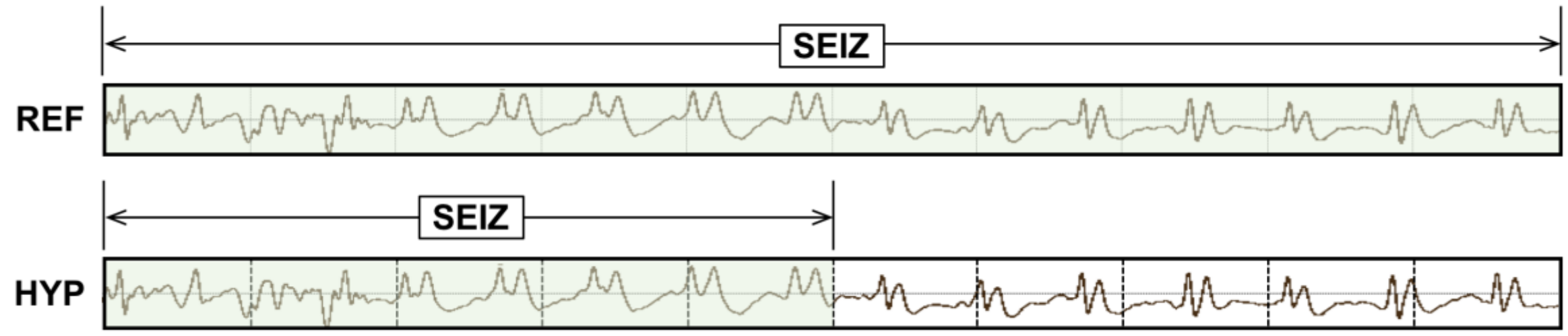

**Figure 1.** A typical situation where a hypothesis (HYP) has a 50% overlap with the reference (REF).

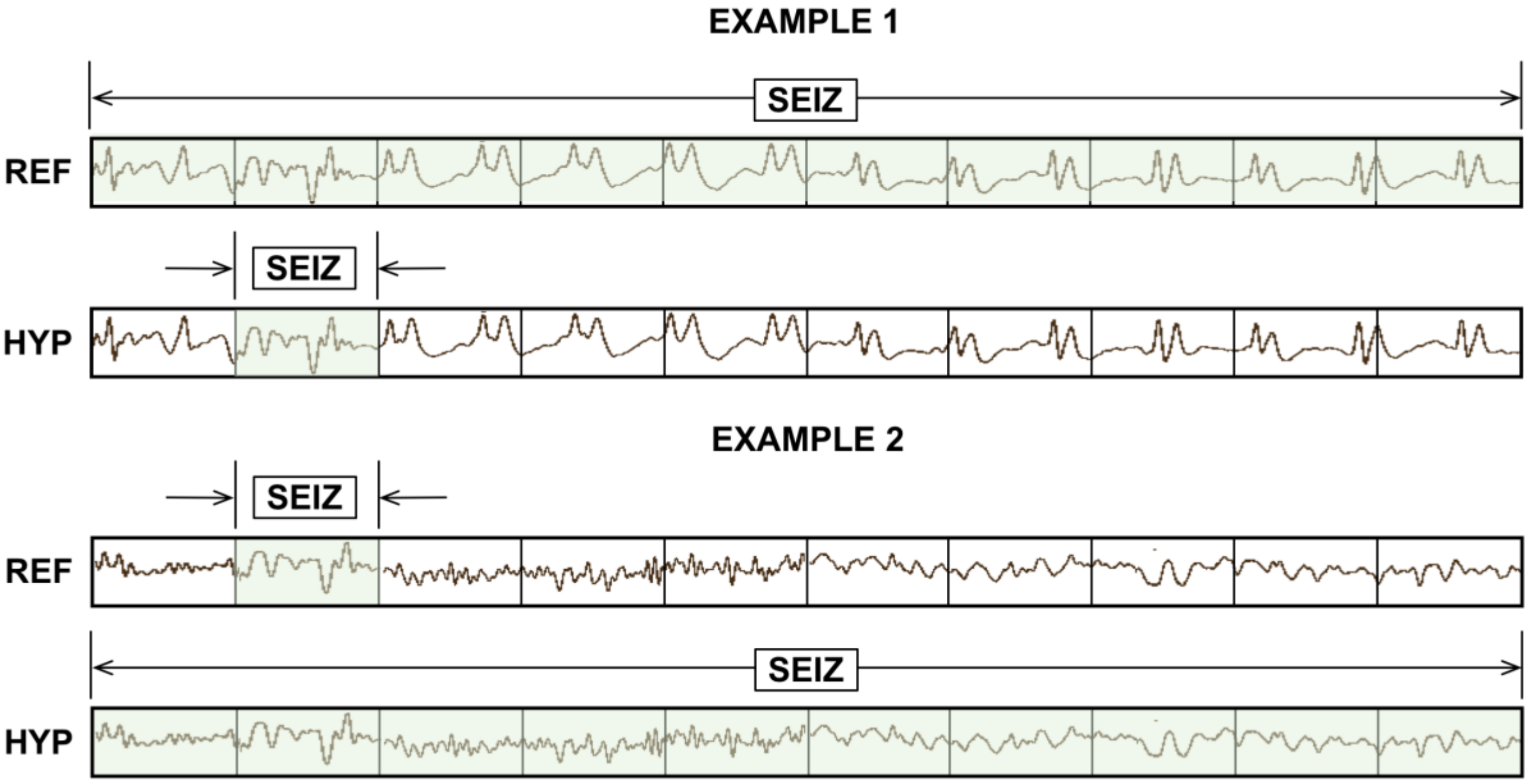

**Figure 2.** TP scores for the Any-Overlap method are *100%* even though large portions of the event are missed.

The proper balance between sensitivity and FA rate is often application specific and has been studied extensively in a number of research communities. For example, evaluation of voice keyword search technology was carefully studied in the Spoken Term Detection (STD) evaluations conducted by NIST [23][24][34]. These evaluations resulted in the introduction of a single metric, Actual Term-Weighted Value (ATWV) [24], to address concerns about tradeoffs for the different types of errors that occur in voice keyword search systems. Despite being popular in the voice processing community, ATWV has not been used in the bioengineering community.

Therefore, in this paper, we compare and contrast five popular scoring metrics and one derived measure:

(1) *NIST Actual Term-Weighted Value (ATWV):* based on NIST's popular scoring package (F4DE v3.3.1), this metric, originally developed for the NIST 2006 Spoken Term Detection evaluation, uses an objective function that accounts for temporal overlap between the reference and hypothesis using the detection scores assigned by the system.

(2) *Dynamic Programming Alignment* (DPALIGN): similar to the NIST package known as SCLite [35], this metric uses a dynamic programming algorithm to align terms. It is most often used in a mode in which the time alignments produced by the system are ignored.

(3) *Epoch-Based Sampling* (EPOCH): treats the reference and hypothesis as temporal signals, samples each at a fixed epoch duration, and counts errors accordingly.

(4) *Any-Overlap* (OVLP): assesses the overlap in time between a reference and hypothesis event, and counts errors using binary scores for each event.

(5) *Time-Aligned Event Scoring* (TAES): similar to (4), but considers the percentage overlap between the two events and weights errors accordingly.

(6) *Inter-Rater Agreement* (IRA): uses EPOCH scoring to estimate errors, and calculates Cohen's Kappa coefficient [36] using the measured TP, TN, FP and FN.

It is important to understand that each of these measures estimates TP, TN, FP and FN through some sort of error analysis. From these estimated quantities, traditional derived measures such as sensitivity and specificity are computed. As a result, we will see that sensitivity is a function of the underlying metric, and this is why it is important there be community-wide agreement on a specific metric.

We now briefly describe each of these approaches and provide several examples that illustrate their strengths and weaknesses. These examples are drawn on a compressed time-scale for illustrative purposes and were carefully selected because they are indicative of scoring metric problems we have observed in actual evaluation data collected from our algorithm research.

## 2.1. NIST Actual Term-Weighted Value (ATWV)

ATWV is a measure that balances sensitivity and FA rate. ATWV essentially assigns an application-dependent reward to each correct detection and a penalty to each incorrect detection. A perfect system results in an ATWV of *1.0*, while a system with no output results in an ATWV of *0*. It is possible for ATWV to be less than zero if a system is doing very poorly (for example a high FA rate). Experiments in voice keyword search have shown that an ATWV greater than *0.5* typically indicates a promising or usable system for information retrieval by voice applications. We believe a similar range is applicable to EEG analysis.

The metric accepts as input a list of *N*-tuples representing the hypotheses for the system being evaluated. Each of these *N*-tuples consists of a start time, end time and system detection score. These entries are matched to the reference annotations using an objective function that accounts for both temporal overlap between the reference and hypotheses and the detection scores assigned by the system being evaluated. These detection scores are often likelihood or confidence scores [23]. The probabilities of miss and FA errors at a detection threshold $\theta$ are computed using:

$$P_{Miss(kw,\theta)} = 1 - {N_{Correct(kw,\theta)}}/{N_{Ref(kw)}} \ , \tag{1}$$

$$P_{FA(kw,\theta)} = {N_{Spurious(kw,\theta)}}/{N_{NT(kw)}} \ , \tag{2}$$

where $N_{Correct(kw,\theta)}$ is the number of correct detections of terms with a detection score greater than or equal to $\theta$, $N_{Spurious(kw,\theta)}$ is the number of incorrect detections of terms with a detection score greater than or equal to $\theta$, and $N_{NT(kw)}$ is number of non-target trials for the term $kw$ in the data. The number of non-target trials for a term is related to the total duration of source signal in seconds, $T_{Source}$, and is computed as $N_{NT(kw)} = T_{Source} - N_{Ref(kw)}$.

A term-weighted value is then computed that specifies a trade-off of misses and FAs. ATWV is defined as the value of TWV at the system's chosen detection threshold. Using a predefined constant, $\beta$, that was optimized experimentally ($\beta = 999.9$) [24], ATWV is computed using:

$$TWV_{(kw,\theta)} = 1 - P_{Miss(kw,\theta)} - \beta \, P_{FA(kw,\theta)} \ . \tag{3}$$

A standard implementation of this approach is available at [37]. This metric has been widely used throughout the human language technology community for *15* years. This is a very important consideration in standardizing such a metric – researchers are using a common shared software implementation that ensures there are no subtle implementation differences between sites or researchers.

To demonstrate the features of this approach, consider the case shown in Figure 3. The hypothesis for this segment consists of several short seizure events while the reference consists of one long event. The ATWV metric will assign a TP score of *100%* because the midpoint of the first event in the hypothesis annotation is mapped to the long seizure event in the reference annotation. This is somewhat generous given that *50%* of the event was not detected. The remaining *5* events in the hypothesis annotation are counted as false positives. The ATWV metric is relatively insensitive to the duration of the reference event, though the *5* false positives will lower the overall performance of the system. The important issue here is that the hypothesis correctly detected about *70%* of the seizure event, and yet because of the large number of false positives, it will be penalized heavily.

In Figure 4 we demonstrate a similar case in which the metric penalizes the hypothesis for missing three seizure events in the reference. Approximately *50%* of the segment is correctly identified. This type of scoring penalizing repeated events that are part of a larger event in the reference might make sense in an application like voice keyword search because in human language each word hypothesis serves a unique purpose in the overall understanding of the signal. However, for a two-class event detection problem such as seizure detection, such scoring too heavily penalizes the hypothesis for splitting a long event into a series of short events.

## 2.2. Dynamic Programming Alignment (DPALIGN)

The DPALIGN metric essentially performs a minimization of an edit distance (the Levenshtein distance) [12] to map the hypothesis onto the reference. DPALIGN determines the minimum number of edits required to transform the hypothesis string into the reference string. Given two strings, the source string $X = [x_1, x_2, ..., x_n]$ of length $n$, and target string $Y = [y_1, y_2, ..., y_m]$ of length $m$, we define $d_{i,j}$, which is the edit distance between the substring $x_1:x_i$ and $y_1:y_j$, as:

$$d_{i,j} = \begin{cases} d_{i-1,j} + del \\ d_{i,j-1} + ins \\ d_{i-1,j-1} + sub \end{cases}, \tag{4}$$

The quantities being measured here are often referred to as substitution (sub), insertion (ins) and deletion (del) penalties. For this study, these three penalties are assigned equal weights of *1*. A dynamic programming algorithm is used to find the optimal alignment between the reference and hypothesis based on these weights. Though there are versions of this metric that perform time-aligned scoring in which both the reference and hypothesis must include start and end times, this metric is most commonly used without time alignment information.

The metric is best demonstrated using the two examples shown in Figure 5. In the first example, the reference signal had three seizure events but the hypothesis only detected two seizure events, so there were two insertion errors. In the second example the hypothesis missed the third seizure event, so there were two deletion errors. For convenience, lowercase symbols indicate correct detections while uppercase symbols indicate errors. The asterisk symbol is used to denote deletion and insertion errors. Note that there is ambiguity in these alignments. For example, it is not really clear which of the three seizure events in the second example corresponded to each of the seizure events in the hypothesis. Nevertheless, this imprecision doesn't really influence the overall scoring. Though this type of scoring might at first seem highly inaccurate since it ignores time alignments of the hypotheses, it has been surprisingly effective in scoring machine learning systems in sequential data applications (e.g., speech recognition) [12][35].

## 2.3. Epoch-Based Sampling (EPOCH)

Epoch-based scoring uses a metric that treats the reference and hypothesis as signals. These signals are sampled at a fixed epoch duration. The corresponding label in the reference is compared to the hypothesis. Similar to DPALIGN, substititions, deletions and insertion errors are tabulated with an equal weight of *1* for each type of error. This process is depicted in Figure 6. Epoch-based scoring requires that the entire signal be annotated, which is normally the case for sequential decoding evaluations. It attempts to account for the amount of time the two annotations overlap, so it directly addresses the inconsistencies demonstrated in Figure 3 and Figure 4.

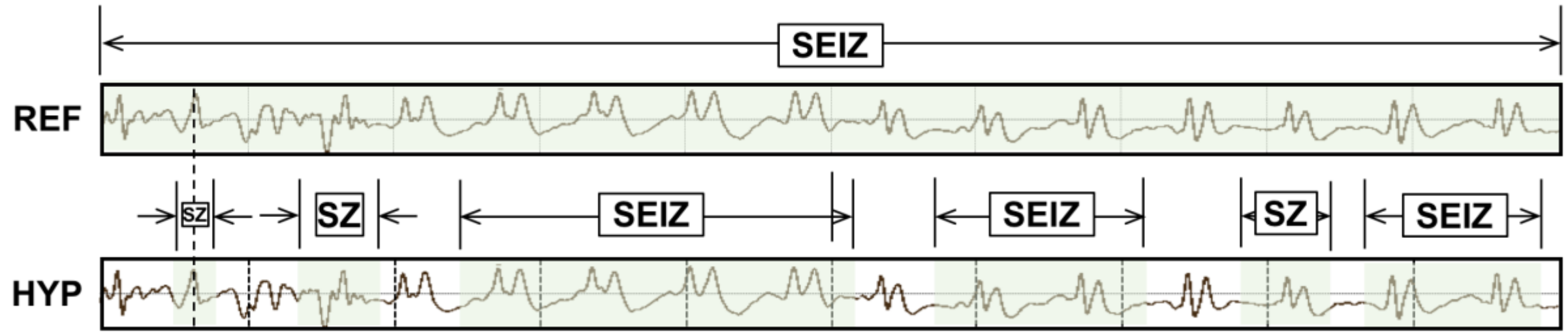

**Figure 3.** ATWV scores this segment as *1* TP and *5* FPs.

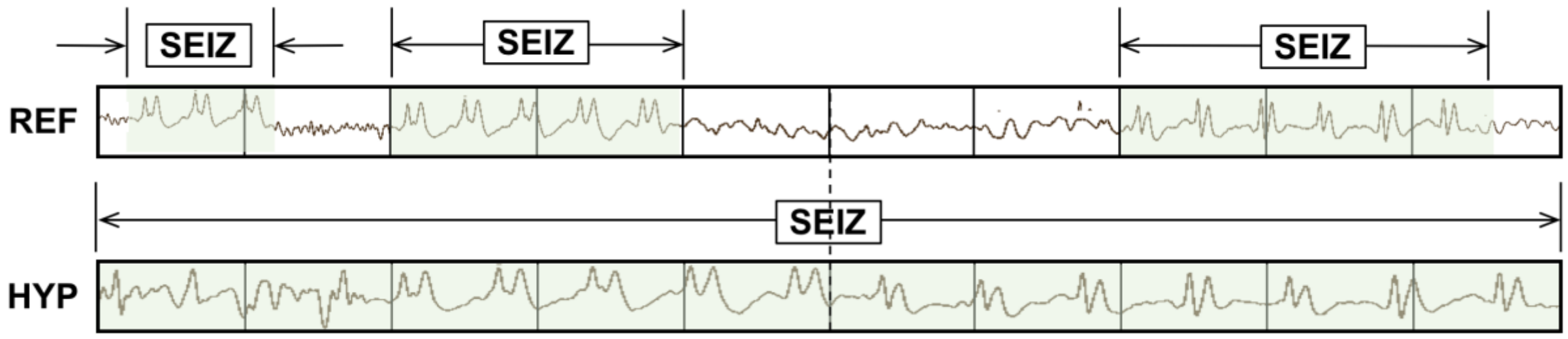

**Figure 4.** ATWV scores this segment as *0* TP and *4* FN events.

```
   Ref: bckg seiz bckg seiz bckg **** ****
   Hyp: bckg seiz bckg seiz bckg SEIZ BCKG
(Hits: 5 Sub: 0 Ins: 2 Del: 0 Total Errors: 2)

   Ref: bckg seiz bckg seiz bckg SEIZ BCKG
   Hyp: bckg seiz bckg seiz bckg **** ****
(Hits: 5 Sub: 0 Ins: 0 Del: 2 Total Errors: 2)
```

**Figure 5.** DPALIGN aligns symbol sequences based on edit distance and ignores time alignments.

One important parameter to be tweaked in this algorithm is the frequency with which we sample the two annotations, which we refer to as the scoring epoch duration. It is ideally set to an amount of time smaller than the unit of time used by the classification system to make decisions. For example, the hypothesis in Figure 6 outputs decisions every *1* sec. The scoring epoch duration should be set smaller than this. We use a scoring epoch duration of *0.25* sec for most of our work because our system epoch duration is typically *1* sec. We find in situations like this the results are not overly sensitive to the choice of the epoch duration as long as it is below *1* sec. This parameter simply controls how much precision one expects for segment boundaries.

Because EPOCH scoring samples the annotations at fixed time intervals, it is inherently biased to weigh long seizure events more heavily. For example, if a signal contains one extremely long seizure event (e.g., *1000* secs) and two short events (e.g., each *10* secs in duration), the accuracy with which the first event is detected will dominate the overall scoring. Since seizure events can vary dramatically in duration, this is a cause for concern.

## 2.4. Any-Overlap Method (OVLP)

We previously introduced the OVLP metric as a popular choice in the neuroengineering community [30][31]. OVLP is a more permissive metric that tends to produce much higher sensitivities. If an event is detected in close proximity to a reference event, the reference event is considered correctly detected. If a long event in the reference annotation is detected as multiple shorter events in the hypothesis, the reference event is also considered correctly detected. Multiple events in the hypothesis annotation corresponding to the same event in the reference annotation are not typically counted as FAs. Since the FA rate is a very important measure of performance in critical care applications, this is another cause for concern.

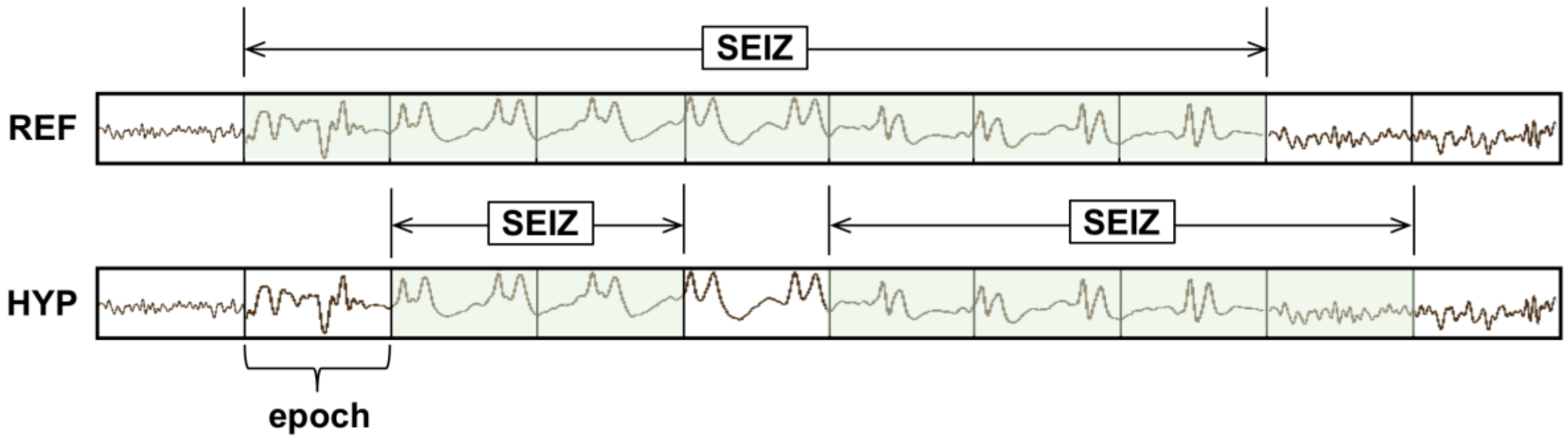

**Figure 6.** EPOCH scoring directly measures the similarity of the time-aligned annotations. The TP, FN and FP are *5*, 2 and *1* respectively

The OVLP scoring method is demonstrated in Figure 7. It has one significant tunable parameter – a guard band that controls the degree to which a misalignment is still considered as a correct match. In this study, we use a fairly strict interpretation of this band and require some overlap between the two events in time – essentially a guard band of zero. The guard band needs to be tuned based on the needs of the application. Sensitivity generally increases as the guard band is increased.

## 2.5. Time-Aligned Event Scoring (TAES)

Though EPOCH scoring directly measures the amount of overlap between the annotations, there is a possibility that this too heavily weights single long events. Seizure events can vary in duration from a few seconds to many hours. In some applications, correctly detecting the number of events is as important as their duration. Hence, the TAES metric was designed as a compromise between these competing constraints. The TAES scoring metri calculates TP and FP as follows:

$$TP_{(term)} = {Duration_{correct(term)}}\Big/{Duration_{Ref(term)}} \tag{5}$$

$$Duration_{correct(term)} = \begin{cases} H_{stop(tetm)} - H_{start(term)}, & if\ \ H_{start(term)} \geq R_{strat}(term),\ H_{stop(term)} \leq R_{stop(term)} \\ H_{stop(tetm)} - R_{start(term)}, & if\ \ H_{start(term)} < R_{start(term)},\ H_{stop(term)} \leq R_{stop(term)} \\ R_{stop(tetm)} - H_{start(term)}, & if\ \ H_{start(term)} \geq R_{strat(term)},\ H_{stop(term)} > R_{stop(term)} \\ R_{stop(tetm)} - R_{start(term)}, & otherwisde \end{cases} \tag{6}$$

$$FP_{(term)} = \begin{cases} {Duration_{Spurious(term)}}\Big/{Duration_{Ref(term)}}, & if\ Duration_{Spurious(term)} < Duration_{Ref(term)}, \\ 1, & otherwise, \end{cases} \tag{7}$$

Where $H$ and $R$ represent the reference and hypothesis events respectively, and $Duration_{Ref(term)}$ represents the duration of the reference events. The $Duration_{Spurious(term)}$ is duration of inserted term.

TAES gives equal weight to each event, but it calculates a partial score for each event based on the amount of overlap. The TP score is the total duration of a detected term divided by the total duration of the reference term. The FN score is the fraction of the time the reference term was missed divided by the total duration of the reference term. The FP score is the total duration of the inserted term divided by total amount of time this inserted term was incorrect according to the reference annotation. But FP can't be more than 1 per event. Therefore, like TP and FN, a single FP event contributes a fractional amount to the overall FP score if it correctly detects a portion of the same event in the reference annotation (partial overlap). Moreover, if multiple reference events are detected by a single long hypothesis event, all but the first detections are considered as FNs. Since, FP per event cannot exceed 1, this property helps compensating the sensitivity versus FA trade-off. An example of TAES scoring is depicted in Figure 8.

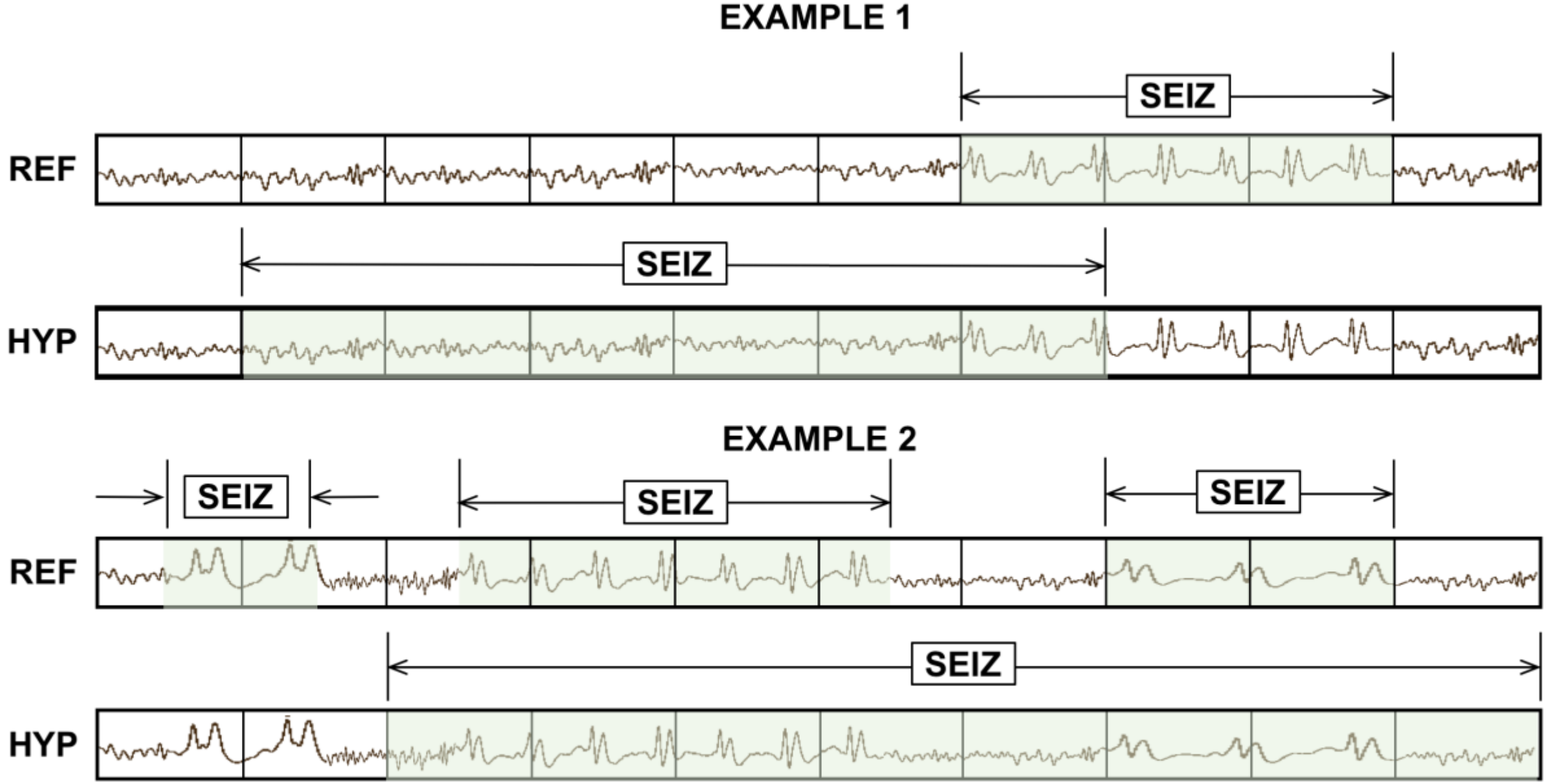

**Figure 7.** OVLP scoring is very permissive about the degree of overlap between the reference and hypothesis. The TP score for example 1 is *1* with no false alarms. In example 2, the system detects *2* out of *3* seizure events, so the TP and FN scores are *2* and *1* respectively.

### 2.6. Inter-Rater Agreement (IRA)

Inter-rater agreement (IRA) is a popular measure when comparing the relative similarity of two annotations. We refer to this metric as a derived metric since it is computed from error counts collected using one of the other five metrics. IRA is most often measured using Cohen's Kappa coefficient [36], which compares the observed accuracy with the expected accuracy. It is computed using:

$$\kappa = \frac{p_0 - p_e}{1 - p_e}, \qquad (8)$$

where $p_o$ is the relative observed agreement among raters and $p_e$ is the hypothetical probability of chance agreement.

The Kappa coefficient ranges between $\kappa = 1$ (complete agreement) and $-1 \leq \kappa \leq 0$ (no agreement). It has been used extensively to assess inter-rater agreement for experts manually annotating seizures in EEG signals. Values in the range of $0.5 \leq \kappa \leq 0.8$ are common for these types of assessments [38]. The variability amongst experts mainly involves fine details in the annotations, such as the exact onset of a seizure. These kinds of details are extremely important for machine learning and hence we need a metric that is sensitive to small variations in the annotations. For completeness, we use this measure as a way of evaluating the amount of agreement between two annotations.

### 2.7. A Brief Comparison of Metrics

A simple example of how these metrics compare on a specific segment of a signal is shown in Figure 9. A *10*-sec section of an EEG signal is shown subdivided into *1*-sec segments. The reference has three isolated events. The system being evaluated outputs one hypothesis that starts in the middle of the first event and continues through the remaining two events. ATWV scores the system as *1* TP and *2* FNs since it assigns the extended hypothesis event to the center reference event and leaves the other two undetected. The ATWV score is *0.33* for seizure events, *0.25* for background events, resulting in an average ATWV of *0.29*. The sensitivity and FA rates for seizure events for this metric are *33%* and *0* per *24* hrs. respectively. DPALIGN scores the system the same way since time alignments are ignored and the first event in each annotation are matched together, leaving the other two events undetected.

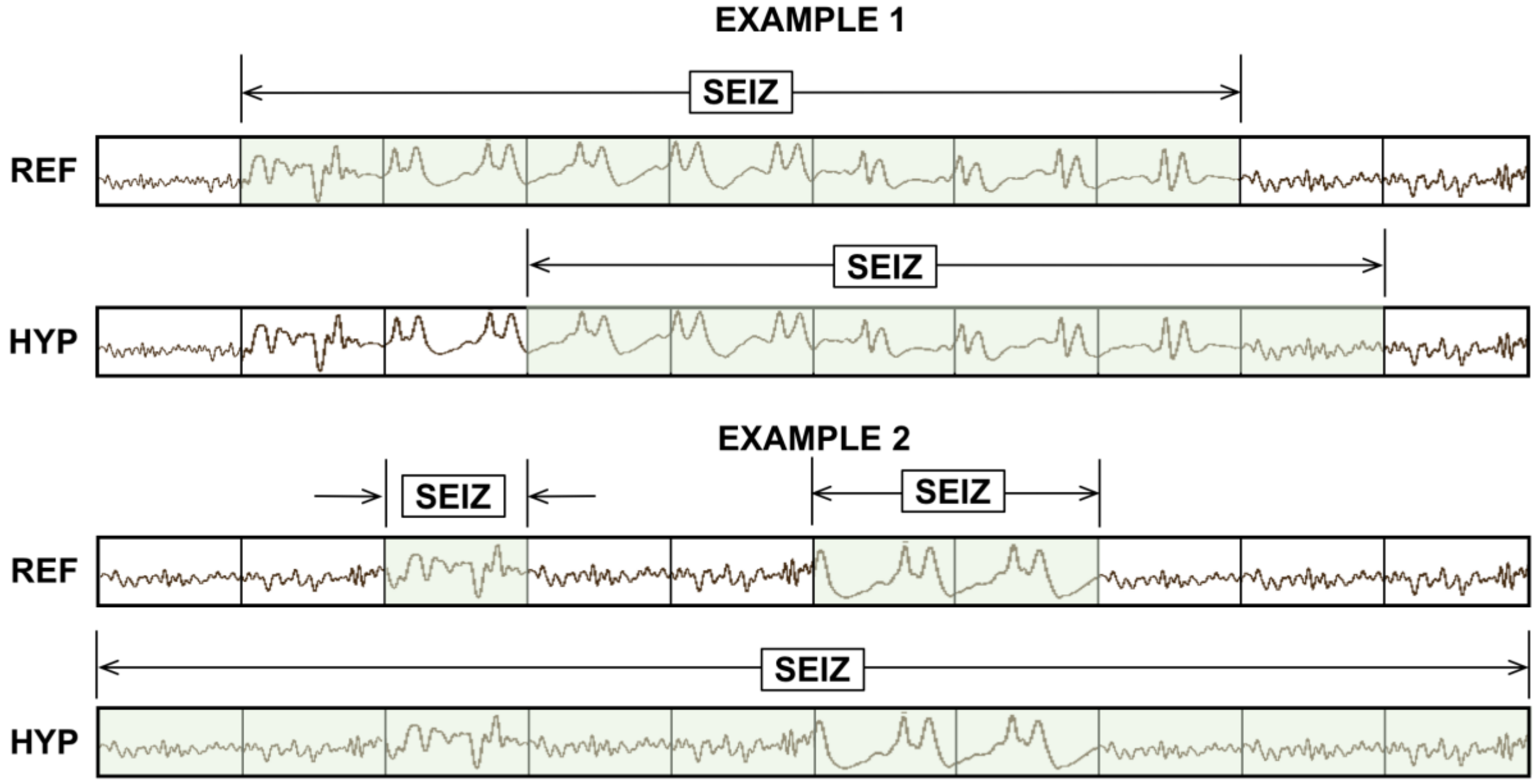

**Figure 8.** TAES scoring accounts for the amount of overlap between the reference and hypothesis. TAES scores example 1 as 0.71 TP, 0.29 FN and 0.14 FP. Example 2 is scored as 1 TP, 1 FN and 1 FP.

The EPOCH method scores the alignment *5* TP, *3* FP and *1* FN using a *1*-sec epoch duration because there are *4* epochs for which the annotations do not agree and *5* epochs where they agree. The sensitivity is *83.33%* and the FA rate per 24 hrs. is very high because of the *3* FPs. The OVLP method scores the segment as *3* TP and *0* FP because detected events have partial to full overlap with all the reference events, giving a sensitivity of *100%* with an FA rate of *0*. TAES scores this segment as *0.5* FN and *2.5* TP because the first event is only *50%* correct and there are a TP for the $5^{th}$ to $8^{th}$ and $10^{th}$ epochs (multiple overlapping reference events), giving a sensitivity of *83.33%* and a high FA rate.

IRA for seizure events evaluated using Cohen's Kappa statistic is *0.09* because there are essentially *4* errors for *6* seizure events. IRAs below *0.5* indicate a poor match between the reference and the hypothesis.

It is difficult to conclude from this example which of these measures are most appropriate for EEG analysis. However, we see that ATWV and DPALIGN generally produce similar results. The EPOCH metric produces larger counts because it samples time rather than events. OVLP produces a high sensitivity while TAES produces a low sensitivity but a relatively higher FA rate.

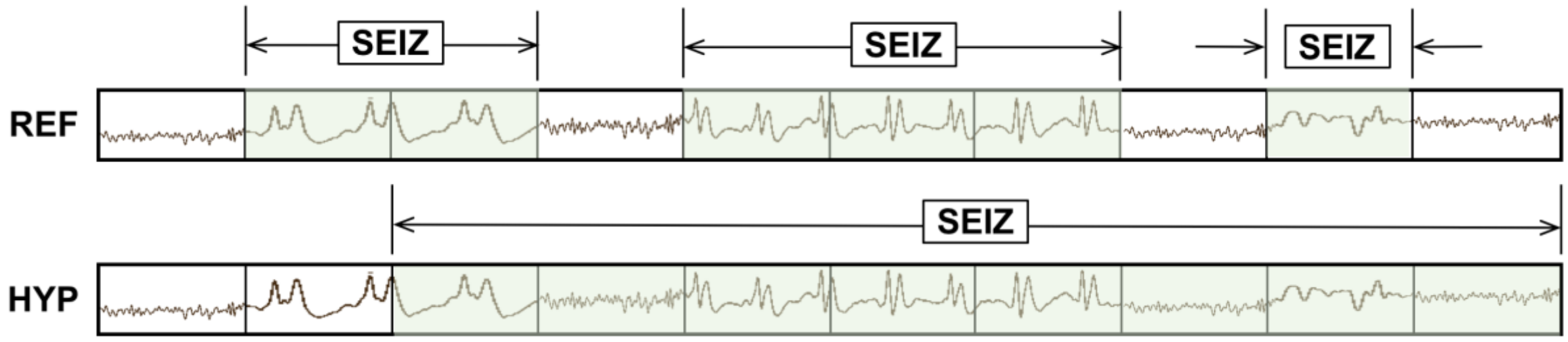

**Figure 9**. An example that summarizes the differences between scoring metrics.

## 3. Results

To demonstrate the differences between these metrics on a realistic task, we have evaluated a range of machine learning systems on a seizure detection task based on the TUH EEG Seizure Corpus [39]. An overview of the corpus is given in Table 1. This is the largest open source corpus of its type. It consists of clinical data collected at Temple University Hospital, and represents a very challenging machine learning task because it contains a rich variety of common real-world problems found in clinical data (e.g., patient movement). There are *50* patients in the evaluation corpus, making it large enough to accurately assess fine differences in algorithm performance.

**Table 1.** The TUH EEG Seizure Corpus (v1.1.1)

| Description | Train | Eval |
|---|---|---|
| Patients | 196 | 50 |
| Sessions | 456 | 230 |
| Files | 1,505 | 984 |
| No. Seizure Events | 870 | 614 |
| Seizure (secs) | 51,140 | 53,930 |
| Non-Seizure (secs) | 877,821 | 547,728 |
| Total (secs) | 928,962 | 601,659 |

A general architecture for the five machine learning systems evaluated is shown in Figure 10. An EEG signal is input using a European Data Format (EDF) file. The signal is converted to a sequence of feature vectors. A group of frames are classified into an event on a per-channel basis using combination of deep learning networks. The deep learning system essentially looks across multiple epochs, which we refer to as the temporal context, and multiple channels, which we refer to as the spatial context since each channel is associated with a location of an electrode on a patient's head. There are a wide variety of algorithms that can be used to produce a decision from these inputs. Even though seizures occur on a subset of the channels input to such a system, we focus on a single decision made across all channels at each point in time.

The five systems selected were carefully chosen because they represent a range of performance that is representative of state of the art on this task and because these systems exhibit different error modalities on this task. The performance of these systems is sufficiently close so that the impact of these different scoring metrics becomes apparent. The systems selected were:

(1) *HMM/SdA:* a hybrid system consisting of a hidden Markov model (HMM) decoder and a postprocessor that uses a Stacked Denoising Autoencoder (SdA). An *N*-channel EEG was transformed into *N* independent feature streams using a standard sliding window based approach. The hypotheses generated by the HMMs were postprocessed using a second stage of processing that examines the temporal and spatial context. We apply a third pass of postprocessing that uses a stochastic language model to smooth hypotheses involving sequences of events so that we can suppress spurious outputs. This third stage of postprocessing provides a moderate reduction in the false alarm rate.

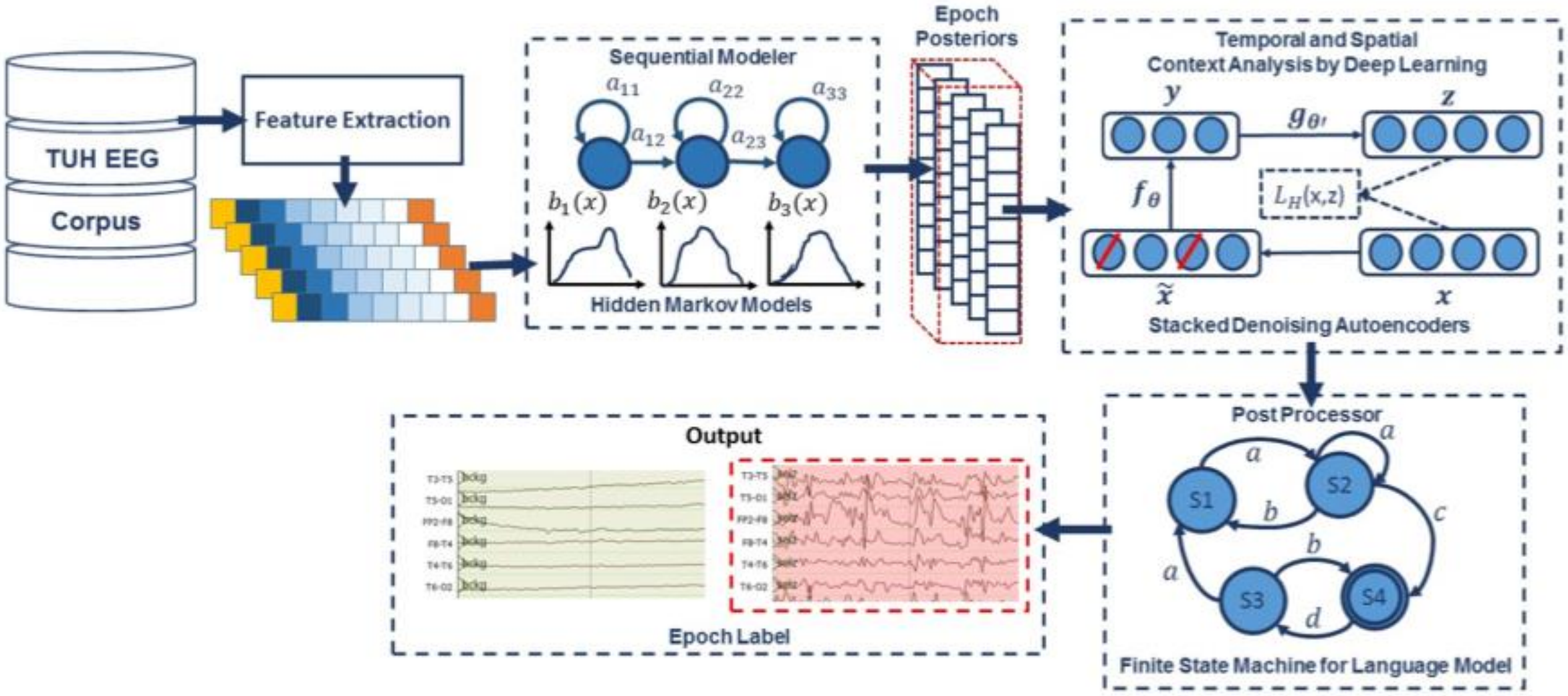

**Figure 10.** A hybrid deep learning architecture that integrates temporal and spatial context

Standard three state left-to-right HMMs with *8* Gaussian mixture components per state were used for sequential decoding. We divide each channel of an EEG into *1*-second epochs, and further subdivide these epochs into a sequence of frames. Each epoch is classified using an HMM trained on the subdivided epoch, and then these epoch-based decisions are postprocessed by additional statistical models in a process similar to the language modeling component of a speech recognizer.

The output of the epoch-based decisions was postprocessed by a deep learning system. The SdA network has three hidden layers with corruption levels of *0.3* for each layer. The number of nodes per layer are: first layer = *800*, second layer = *500*, third layer = *300*. The parameters for pre-training are: learning rate = *0.5*, number of epochs = *150*, batch size = *300*. The parameters for fine-tuning are: learning rate = *0.1*, number of epochs = *300*, batch size = *100*. The overall result of the second stage is a probability vector of dimension two containing a likelihood that each label could have occurred in the epoch. A soft decision paradigm is used rather than a hard decision paradigm because this output is smoothed in the third stage of processing.

(2) *HMM/LSTM:* an HMM decoder postprocessed by a Long Short-Term Memory (LSTM) network. Like the HMM/SdA hybrid approach previously described, the output of the HMM system is a vector of dimension $2 \times$ number of channels (*22*) $\times$ the window length (*7*). Therefore, we also use PCA before LSTM in this approach to reduce the dimensionality of the data to *20*. For this study, we used a window length of *41* for LSTM, and this layer is composed of one hidden layer with 32 nodes. The output layer nodes in this LSTM level use a sigmoid function. The parameters of the models are optimized to minimize the error using a cross-entropy loss function. Adaptive Moment Estimation (Adam) is used in the optimization process.

(3) IPCA/LSTM: a preprocessor based on Incremental Principal Component Analysis (IPCA) followed by an LSTM decoder. The EEG features are delivered to an IPCA layer for spatial context analysis and dimensionality reduction. A batch size of *50* is used in IPCA and the output dimension is *25*. The output of IPCA is delivered to a LSTM for classification. We used a one-layer LSTM with a hidden layer size of *128* and batch size of *128* is used along with Adam optimization and a cross–entropy loss function.

(4) CNN/MLP: a pure deep learning-based approach that uses a Convolutional Neural Network (CNN) decoder and a Multi-Layer Perceptron (MLP) postprocessor. The network contains six convolutional layers, three max pooling layers and two fully-connected layers. A rectified linear unit (ReLU) non-linearity is applied to the output of every convolutional and fully-connected layer.

(5) CNN/LSTM: a pure deep learning-based architecture that uses a combination of CNN and LSTM networks. In this architecture, we integrate 2D CNNs, 1D CNNs and LSTM networks to better exploit long-term dependencies. Exponential Linear Units (ELU) are used as the activation functions for the hidden layers. Adam is used in the optimization process along with a mean squared error loss function.

Comprehensive details about the architectures are available in [40][41]. The details of these systems are not critical to this study. What is more important is how the range of performance is reflected in these metrics.

A comparison of the performance of the different architectures is presented in Table 2. Though the relative rankings of these systems not surprisingly vary with the metric, the ranking of these systems is accurately represented by the overall trends in Table 2. HMM/SdA generally performs the poorest of these systems, delivering a respectable sensitivity but at a high FA rate. CNN/LSTM typically delivers highest performance and has a low FA rate, which is very important in this type of application.

## 4. Discussion

Evaluating systems from a single operating point is always a bit tenuous. Therefore, in Figure 11, we provide DET curves for the systems and in Table 3 we provide AUCs for these DET curves calculated using OVLP and TAES for comparison. This is due to our emphasis on using OVLP and TAES metrics for seizure detection-like applications. The DET curves were derived from output from OVLP scoring metric only.

**Table 2.** Performance vs. scoring metric

| Metric | Measure | HMM/SdA | HMM/LSTM | IPCA/LSTM | CNN/MLP | CNN/LSTM |
|---|---|---|---|---|---|---|
| **ATWV** | Sensitivity | 30.35% | 26.73% | 24.73% | 29.52% | 30.34% |
| | Specificity | 61.38% | 68.93% | 64.51% | 65.87% | 93.15% |
| | FAs/24 hrs | 98 | 75 | 94 | 94 | 11 |
| | ATWV | -0.8392 | -0.8469 | -0.4628 | -0.7971 | 0.1737 |
| **OVLP** | Sensitivity | 35.35% | 30.05% | 32.97% | 39.09% | 30.83% |
| | Specificity | 73.35% | 80.53% | 77.57% | 76.84% | 96.86% |
| | FAs/24 hrs | 77 | 60 | 73 | 77 | 7 |
| **DPALIGN** | Sensitivity | 44.11% | 33.77% | 35.77% | 43.35% | 32.46% |
| | Specificity | 66.87% | 72.99% | 69.59% | 71.49% | 95.17% |
| | FAs/24 hrs | 86 | 66 | 81 | 77 | 8 |
| **TAES** | Sensitivity | 17.29% | 22.84% | 22.12% | 31.58% | 12.48% |
| | Specificity | 66.04% | 70.41% | 66.64% | 64.75% | 95.24% |
| | FAs/24 hrs | 82 | 68 | 83 | 91 | 8 |
| **EPOCH** | Sensitivity | 20.71% | 50.46% | 51.02% | 65.03% | 9.784% |
| | Specificity | 98.22% | 94.82% | 94.09 | 91.55% | 99.84% |
| | FAs/24 hrs | 1418 | 4133 | 4711 | 6738 | 126 |

The shapes of the DET curves do not change significantly with the scoring metric though the absolute numbers vary similarly to what we see in Table 2. AUC values from Table 3 also follows the similar trend but the AUC-TAES difference between the best and the worst system is less pronounced compared to the AUC-OVLP which seems to provide more realistic insight of the system's peroformance. It is clear from this data that CNN/LSTM performance is significantly different from the other systems. This is primarily because of its low FA rate. For this particular application, sensitivity drops rapidly as the FA rate is lowered. Therefore, comparing a single data point for each system is dangerous because the systems are most likely operating at different points on a DET curve if the sensitivities are significantly different. We find tuning these systems to have a comparable FA rate is important when comparing two systems only based on sensitivity.

In Table 2 we can examine the sensitivity of the different metrics by looking at the variation in sensitivity. For example, for HMM/SdA, we see the lowest sensitivities are produced by TAES and EPOCH scoring, while the highest sensitivities are produced by OVLP and DPALIGN. This makes sense because OVLP and DPALIGN are very forgiving of time alignment errors, while TAES and EPOCH penalize time alignment errors heavily. We see similar trends for CNN/LSTM though the range of differences between the three highest scoring metrics is smaller. We also see that the five algorithms are ranked similarly by each scoring metric. HMM/SdA consistently scores the lowest and CNN/LSTM consistently scores the highest. The other three systems are very similar in their performance.

The ATWV scores for all algorithms are extremely low. The ATWV scores are below *0.5* which indicates that overall performance is poor. However, the ATWV score for CNN/LSTM is significantly higher than the other four systems. ATWV attempts to reduce the information contained in a DET curve to a single number, and does a good job reflecting the results shown in Figure 11. The DET curves for HMM/LSTM and HMM/SdA overlap considerably for an FP rate between *0.25* and *1.0*, and this is a primary reason why their ATWV scores are similar. However, for the seizure detection application we are primarily interested in the low FP rate region, and in that range, HMM/LSTM and IPCA/LSTM perform similarly.

**Table 3.** AUC comparison according to OVLP and TAES

| Algorithm | AUC (OVLP) | AUC (TAES) |
|---|---|---|
| HMM/SdA | 0.44 | 0.72 |
| HMM/LSTM | 0.44 | 0.71 |
| IPCA/LSTM | 0.39 | 0.72 |
| CNN/MLP | 0.38 | 0.65 |
| CNN/LSTM | 0.21 | 0.56 |

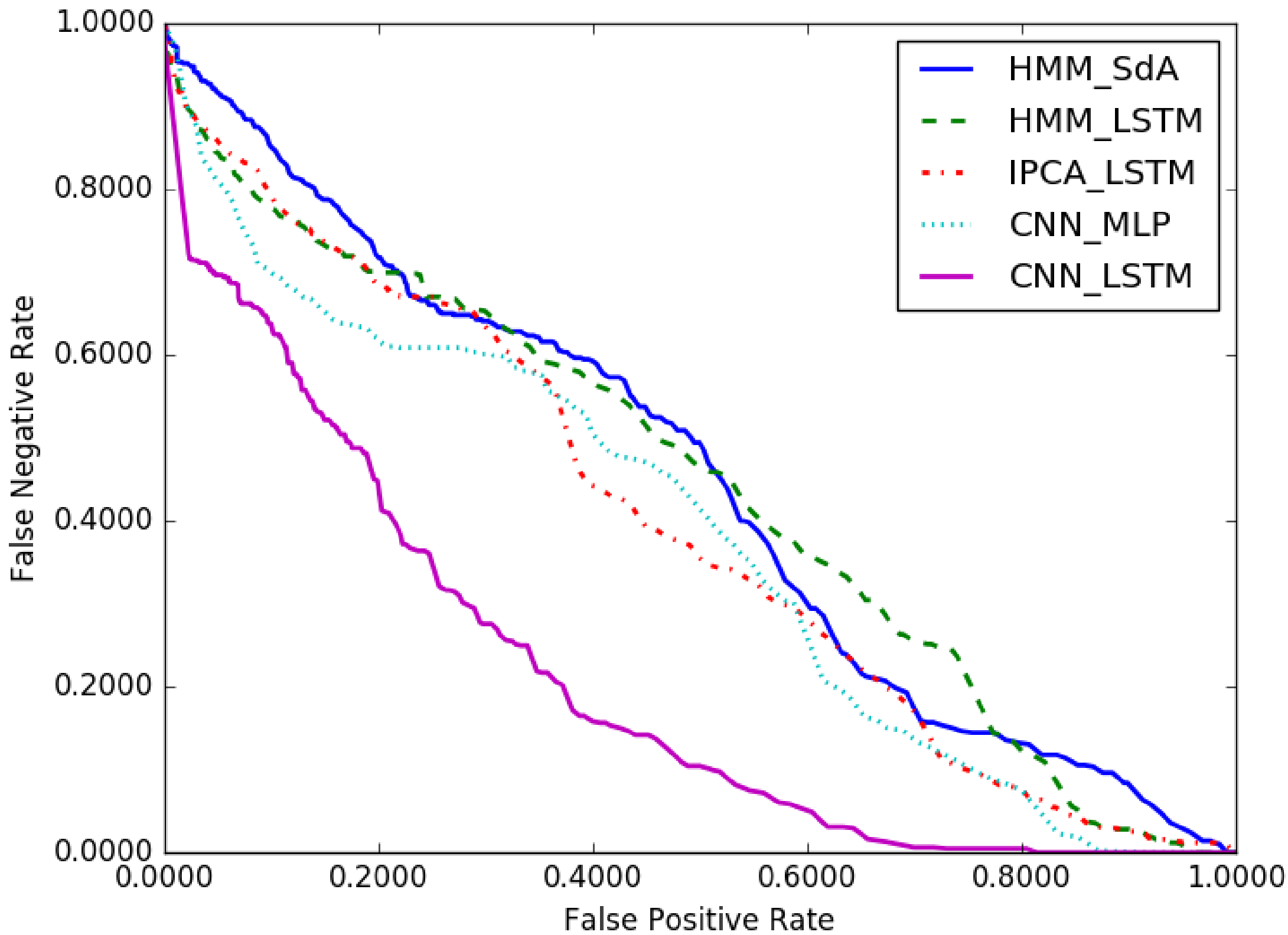

**Figure 11.** A comparison of DET curves

While sensitivity and specificity are commonly used metrics in the bioengineering community, from Table 2 and Figure 11 we see that the FA rate also plays a major role in determining the usability of a system. A commonly used metric in the machine learning community that is somewhat intuitive is accuracy. The accuracy of the five systems is shown in Table 4. Accuracy weights all types of errors as equally important. This is acceptable if the dataset is balanced. However, for many bioengineering applications, such as seizure detection, the target class, or class of interest, occurs infrequently. We see that CNN/LSTM is significantly more accurate than the other four systems, but that the differences between these remaining four systems is minimal when using accuracy as a metric.

Another popular metric that attempts to aggregate performance into a single data point, and is popular in the information retrieval communities, is the $F_1$ score. These scores for the five systems are shown in Table 5. We see there is significant variation in $F_1$ scores with the scoring metric. For example, for TAES and EPOCH, which stress time alignments, the best performing system is not CNN/LSTM. $F_1$ scores do not adequately emphasize FAs for applications such as seizure detection.

We generally prefer operating points where performance in terms of sensitivity, specificity and FAs is balanced. The ATWV metric explicitly attempts to balance these by assigning a reward to each correct detection and a penalty to each incorrect detection. None of the conventional metrics described here consider the fraction of a detected event that is correct. This is the inspiration behind the development of TAES scoring. TAES scoring requires the time alignments to match, which is a more stringent requirement than, for example, OVLP. Consequently, the sensitivity produced by the TAES and EPOCH metrics tends to be lower.

Finally, comparing results across these five metrics can provide useful diagnostic information and provide insight into the system's behavior. For example, the IPCA/LSTM and HMM/LSTM systems have relatively higher sensitivities according to the EPOCH metric, indicating that these systems tend to detect longer seizure events. Conversely, since the CNN/LSTM system has relatively low sensitivities according to the TAES and EPOCH metrics, it can be inferred that this system misses longer seizure events. Similarly, if the sensitivity was relatively high for TAES and relatively low for EPOCH, it would indicate that the system tends to detect a majority of smaller to moderate events precisely regardless of the duration of an event. Similarly, a comparison of ATWV scores with other metrics gives diagnostic information such as whether a system accurately detects the onset and end of an event or whether the system splits long events into multiple short events. Examining the ensemble of scores can be revealing for these six metrics.

**Table 4.** Accuracy vs. scoring metric

| Metric | HMM/SdA | HMM/LSTM | IPCA/LSTM | CNN/MLP | CNN/LSTM |
|---|---|---|---|---|---|
| ATWV | 54.0% | 54.0% | 52.1% | 54.9% | 70.7% |
| OVLP | 65.1% | 66.5% | 65.6% | 66.9% | 78.9% |
| DPALIGN | 61.5% | 60.2% | 59.2% | 62.9% | 73.6% |
| TAES | 56.6% | 57.3% | 55.4% | 57.2% | 69.7% |
| EPOCH | 92.3% | 91.5% | 90.8 % | 89.5% | 91.5% |

**Table 5.** $F_1$ score vs. scoring metric

| Metric | HMM/ SdA | HMM/LSTM | IPCA/LSTM | CNN/MLP | CNN/LSTM |
|---|---|---|---|---|---|
| ATWV | 0.24 | 0.28 | 0.24 | 0.28 | 0.42 |
| OVLP | 0.31 | 0.33 | 0.34 | 0.38 | 0.45 |
| DPALIGN | 0.35 | 0.36 | 0.35 | 0.42 | 0.45 |
| TAES | 0.16 | 0.26 | 0.24 | 0.31 | 0.19 |
| EPOCH | 0.29 | 0.47 | 0.46 | 0.49 | 0.14 |

To understand the pairwise statistical difference between the discussed evaluation metrics and deep architectures, we have performed Kolmogorov-Smirnov (KS), Pearson's R (correlation coefficient) and Z-test. These tests were performed to evaluate results of hybrid deep learning architectures on the basis of sensitivity and specificity. Each individual patient from the TUSZ dataset was evaluated separately. Outliers were removed by rejecting all input values collected from patients which have no seizures and from those for which deep learning systems detected no seizures.

Prior to performing tests for evaluating statistically differences, such as a z-test, *t*-test or ANOVA, it must first be determined whether or not the group sample, in our case individual metric's score on per patient evaluation, is normally distributed. We performed KS tests on each separate evaluation metric and confirmed that the group distribution is indeed Gaussian. The KS values for normal distributions collected range from *0.61 – 0.71* for sensitivity and *0.99 – 1.00* for specificity with the *p*-values equal to zero. We then evaluate the correlation coefficient (Pearson's R) between each metric-pairs.

**Table 6.** Correlation of the scoring metrics (for sensitivity)

| Metric | ATWV | DPALIGN | OVLP | TAES | EPOCH |
|---|---|---|---|---|---|
| **ATWV** | --- | 0.87 ($p < 0.001$) | 0.92 ($p < 0.001$) | 0.71 ($p < 0.001$) | 0.50 ($p < 0.001$) |
| **DPALIGN** | 0.87 ($p < 0.001$) | --- | 0.90 ($p < 0.001$) | 0.69 ($p < 0.001$) | 0.48 ($p < 0.001$) |
| **OVLP** | 0.92 ($p < 0.001$) | 0.90 ($p < 0.001$) | --- | 0.78 ($p < 0.001$) | 0.62 ($p < 0.001$) |
| **TAES** | 0.71 ($p < 0.001$) | 0.69 ($p < 0.001$) | 0.78 ($p < 0.001$) | --- | 0.87 ($p < 0.001$) |
| **EPOCH** | 0.50 ($p < 0.001$) | 0.48 ($p < 0.001$) | 0.62 ($p < 0.001$) | 0.87 ($p < 0.001$) | --- |

**Table 7.** Correlation of the scoring metrics (for specificity)

| Metric | ATWV | DPALIGN | OVLP | TAES | EPOCH |
|---|---|---|---|---|---|
| **ATWV** | --- | 0.49 ($p < 0.001$) | 0.45 ($p < 0.001$) | 0.54 ($p < 0.001$) | 0.32 ($p < 0.001$) |
| **DPALIGN** | 0.49 ($p < 0.001$) | --- | 0.94 ($p < 0.001$) | 0.89 ($p < 0.001$) | 0.38 ($p < 0.001$) |
| **OVLP** | 0.45 ($p < 0.001$) | 0.94 ($p < 0.001$) | --- | 0.95 ($p < 0.001$) | 0.44 ($p < 0.001$) |
| **TAES** | 0.54 ($p < 0.001$) | 0.89 ($p < 0.001$) | 0.95 ($p < 0.001$) | --- | 0.56 ($p < 0.001$) |
| **EPOCH** | 0.32 ($p < 0.001$) | 0.38 ($p < 0.001$) | 0.44 ($p < 0.001$) | 0.56 ($p < 0.001$) | --- |

**Table 8.** Significance calculated for scoring metrics using Z-tests for α-value 0.05 (For sensitivity)

| **ATWV (Abs. sensitivity difference (%), Significant/Non-significant)** | | | | | |
|---|---|---|---|---|---|
| **ML Systems (Sens.)** | **CNN-LSTM** | **CNN-MLP** | **HMM-LSTM** | **HMM-SDA** | **IPCA-LSTM** |
| **CNN-LSTM(30.34%)** | --- | (00.82%) Y | (03.61%) Y | (00.01%) Y | (05.61%) Y |
| **CNN-MLP(29.52%)** | | --- | (02.79%) N | (00.83%) N | (04.79%) N |
| **HMM-LSTM(26.73%)** | | | --- | (03.62%) N | (02.00%) N |
| **HMM-SDA(30.35%)** | | | | --- | (05.62%) N |
| **IPCA-LSTM(24.73%)** | | | | | --- |
| **DPALIGN (Abs. sensitivity difference)** | | | | | |
| **ML Systems (Sens.)** | **CNN-LSTM** | **CNN-MLP** | **HMM-LSTM** | **HMM-SDA** | **IPCA-LSTM** |
| **CNN-LSTM(32.46%)** | --- | (10.89%) Y | (01.31%) Y | (11.65%) Y | (03.31%) Y |
| **CNN-MLP(43.35%)** | | --- | (09.58%) N | (00.76%) N | (07.58%) N |
| **HMM-LSTM(33.77%)** | | | --- | (10.34%) N | (02.00%) N |
| **HMM-SDA(44.11%)** | | | | --- | (08.34%) N |
| **IPCA-LSTM(35.77%)** | | | | | --- |
| **EPOCH (Abs. sensitivity difference)** | | | | | |
| **ML Systems (Sens.)** | **CNN-LSTM** | **CNN-MLP** | **HMM-LSTM** | **HMM-SDA** | **IPCA-LSTM** |
| **CNN-LSTM(09.78%)** | --- | (55.25%) N | (40.68%) N | (10.93%) Y | (41.24%) N |
| **CNN-MLP(65.03%)** | | --- | (14.57%) Y | (44.32%) Y | (14.01%) N |
| **HMM-LSTM(50.46%)** | | | --- | (29.75%) Y | (00.56%) N |
| **HMM-SDA(20.71%)** | | | | --- | (30.31%) Y |
| **IPCA-LSTM(51.02%)** | | | | | --- |
| **OVLP (Abs. sensitivity difference)** | | | | | |
| **ML Systems (Sens.)** | **CNN-LSTM** | **CNN-MLP** | **HMM-LSTM** | **HMM-SDA** | **IPCA-LSTM** |
| **CNN-LSTM(30.83%)** | --- | (08.26%) Y | (02.14%) Y | (04.52%) Y | (02.14%) Y |
| **CNN-MLP(39.09%)** | | --- | (09.04%) N | (03.74%) N | (06.12%) N |
| **HMM-LSTM(30.05%)** | | | --- | (05.30%) N | (02.92%) N |
| **HMM-SDA(35.35%)** | | | | --- | (02.38%) N |
| **IPCA-LSTM(32.97%)** | | | | | --- |
| **TAES (Abs. sensitivity difference)** | | | | | |
| **ML Systems (Sens.)** | **CNN-LSTM** | **CNN-MLP** | **HMM-LSTM** | **HMM-SDA** | **IPCA-LSTM** |
| **CNN-LSTM(12.48%)** | --- | (19.10%) N | (10.36%) N | (04.81%) Y | (09.64%) N |
| **CNN-MLP(31.58%)** | | --- | (08.74%) N | (14.29%) Y | (09.46%) N |
| **HMM-LSTM(22.84%)** | | | --- | (05.55%) Y | (00.72%) N |
| **HMM-SDA(17.29%)** | | | | --- | (04.83%) Y |

Correlations for each pair of scoring metrics are shown in Table 6 (for sensitivity) and Table 7 (for specificity). From Table 6, it can be seen that the pairs ATWV-EPOCH and DPALIGN-EPOCH, have minimum correlation (~*0.5*). The pairwise correlations between OVLP, ATWV and DPALIGN are much higher. The EPOCH method has a low correlation with all other metrics except TAES metric. This makes sense because the EPOCH method scores events on a constant time scale instead of on individual events. TAES takes into account the duration of the overlap, so it is the closest method to EPOCH in this regard.

Since OVLP and TAES both score overlapping events independently, we also expect these two methods to be correlated (sensitivity: *0.78*; specificity: *0.95*). ATWV on the other hand has fairlow correlations with the other metrics for specificity because of its stringent rules for FPs when there are multiple overlapping events. The overall highest correlation is between ATWV and OVLP for sensitivity, and OVLP and TAES for specificity. All the correlation values (Pearson's R) collected in Table 6 and Table 7 are statistically significant with the $p$-values < $0.001$.

**Table 9.** Significance calculated for scoring metrics using Z-tests for α-value 0.05 (For specificity)

| **ATWV (Abs. specificity difference (%), Significant/Non-significant)** | | | | | |
|---|---|---|---|---|---|
| **ML Systems (Spec.)** | **CNN-LSTM** | **CNN-MLP** | **HMM-LSTM** | **HMM-SDA** | **IPCA-LSTM** |
| **CNN-LSTM(93.15%)** | --- | (27.28%) Y | (24.22%) Y | (31.77%) Y | (28.64%) Y |
| **CNN-MLP(65.87%)** | | --- | (03.06%) N | (04.49%) N | (01.36%) N |
| **HMM-LSTM(68.93%)** | | | --- | (07.55%) Y | (04.42%) N |
| **HMM-SDA(61.38%)** | | | | --- | (03.13%) N |
| **IPCA-LSTM(64.51%)** | | | | | --- |
| **DPALIGN (Abs. specificity difference (%), Significant/Non-significant)** | | | | | |
| **ML Systems (Spec.)** | **CNN-LSTM** | **CNN-MLP** | **HMM-LSTM** | **HMM-SDA** | **IPCA-LSTM** |
| **CNN-LSTM(95.17%)** | --- | (23.68%) Y | (22.18%) Y | (28.30%) Y | (25.58%) Y |
| **CNN-MLP(71.49%)** | | --- | (01.50%) N | (04.62%) Y | (01.90%) N |
| **HMM-LSTM(72.99%)** | | | --- | (06.12%) Y | (03.40%) N |
| **HMM-SDA(66.87%)** | | | | --- | (02.72%) Y |
| **IPCA-LSTM(69.59%)** | | | | | --- |
| **EPOCH (Abs. specificity difference (%), Significant/Non-significant)** | | | | | |
| **ML Systems (Spec.)** | **CNN-LSTM** | **CNN-MLP** | **HMM-LSTM** | **HMM-SDA** | **IPCA-LSTM** |
| **CNN-LSTM(99.84%)** | --- | (08.29%) N | (05.02%) N | (01.62%) N | (05.75%) N |
| **CNN-MLP(91.55%)** | | --- | (03.27%) N | (06.67%) N | (02.54%) N |
| **HMM-LSTM(94.82%)** | | | --- | (03.40%) N | (00.73%) N |
| **HMM-SDA(98.22%)** | | | | --- | (04.13%) N |
| **IPCA-LSTM(94.09%)** | | | | | --- |
| **OVLP (Abs. specificity difference (%), Significant/Non-significant)** | | | | | |
| **ML Systems (Spec.)** | **CNN-LSTM** | **CNN-MLP** | **HMM-LSTM** | **HMM-SDA** | **IPCA-LSTM** |
| **CNN-LSTM(96.86%)** | --- | (20.02%) Y | (16.33%) Y | (23.51%) Y | (19.29%) Y |
| **CNN-MLP(76.84%)** | | --- | (03.69%) N | (03.49%) Y | (00.73%) N |
| **HMM-LSTM(80.53%)** | | | --- | (07.18%) Y | (02.96%) N |
| **HMM-SDA(73.35%)** | | | | --- | (04.22%) Y |
| **IPCA-LSTM(77.57%)** | | | | | --- |
| **TAES (Abs. specificity difference (%), Significant/Non-significant)** | | | | | |
| **ML Systems (Spec.)** | **CNN-LSTM** | **CNN-MLP** | **HMM-LSTM** | **HMM-SDA** | **IPCA-LSTM** |
| **CNN-LSTM(95.24%)** | --- | (31.21%) Y | (24.83%) Y | (29.20%) Y | (28.60%) Y |
| **CNN-MLP(64.03%)** | | --- | (06.38%) N | (02.01%) Y | (02.61%) N |
| **HMM-LSTM(70.41%)** | | | --- | (04.37%) Y | (03.77%) N |
| **HMM-SDA(66.04%)** | | | | --- | (00.60%) Y |
| **IPCA-LSTM(66.64%)** | | | | | --- |

To understand the statistical significance of each system, we perform (two-tailed) Z-tests on all the recognition system pairs using individual metric separately as shown in Table 8 (for sensitivity) and Table 9 (for specificity). Entries in both these tables have the sensitivity/specificity differenacne between the systems and a binary classification value (Yes/No) based on extracted p-values from the Z-test with 95% confidence. Here again, the data was prepared by scoring systems on individual patients and prior to performing Z-tests, Gaussianity of each sample was evaluated using KS-test. All the samples were confirmed as normal with the p-values < 0.001.

From Table 8, it can be observed that the difference between CNN-LSTM system and other sytems are statistically significant for all metrices except EPOCH and TAES metrices. On the other hand, fails to reject EPOCH and TAES metrices fail to reject null-hypothesis for CNN-LSTM. According to these metric's, the performance of HMM-SDA shows significant difference from other systems showing very poor performance. This can also be observed from EPOCH/TAES results of Table 2.

Table 9 for specificity, shows a different trend than for the sensitivity where EPOCH fails to reject null-hypothesis for all the systems. Since, the specificity is calculated from TN and FP values, for the evalset of duration ~167 hours with epoch size 0.25, few thousand seconds of FPs do not make any significant difference in terms of specificity. This can also be directly observed in Table 2 where specificity of all the systems according to EPOCH is always greater than 90%. The huge difference between the duration of background and seizure events is the primary reason for such high specificities. On the other hand, the OVLP and TAES completely agrees with each other's Z-test results for specificity.

## 5. Conclusions

Standardization of scoring metrics is an extremely important step for a research community to take in order to make progress on machine learning problems such as automatic interpretation of EEGs. There has been a lack of standardization in most bioengineering fields. Popular metrics such as sensitivity and specificity do not completely characterize the problem and neglect the importance that FA rate plays in achieving clinically acceptable solutions. In this paper, we have compared several popular scoring metrics and demonstrated the value of considering the accuracy of time alignments in the overall assessment of a system. We have proposed the use of a new metric, TAES scoring, which is consistent with popular scoring approaches such as OVLP, but provides more accurate assessments by producing fractional scores for recognition of events based on the degree of match in the time alignments. We have also demonstrated the efficacy of an existing metric, ATWV, that is popular in the speech recognition community.

We have also not discussed the extent to which we can tune these metrics by weighting various types of errors based on feedback from clinicians and other 'customers' of the technology. Optimization of the metric is a research problem in itself, since many considerations, including usability of the technology and a broad range of applications, must be involved in this process. Our informal attempts to optimize ATWV and OVLP for seizure detection have not yet produced significantly different results than what was presented here. Feedback from clinicians has been consistent that FA rate is perhaps the single most important measure once sensitivity is above approximately *75%*. As we move more technology into operational environments we expect to have more to contribute to this research topic.

Finally, the Python implementation of these metrics is available at the project web site: *https://www.isip.piconepress.com/projects/tuh_eeg/downloads/nedc_eval_eeg*. Readers are encouraged to refer to the software for detailed questions about the specific implementations of these algorithms and the tunable parameters available.

**Acknowledgments**

Research reported in this publication was most recently supported by the National Human Genome Research Institute of the National Institutes of Health under award number U01HG008468. The content is solely the responsibility of the authors and does not necessarily represent the official views of the National Institutes of Health. This material is also based in part upon work supported by the National Science Foundation under Grant No. IIP-1622765. Any opinions, findings, and conclusions or recommendations expressed in this material are those of the author(s) and do not necessarily reflect the views of the National Science Foundation.